\documentclass[preprint,review, 12pt]{elsarticle}

\journal{Displays}

\usepackage{color}
\usepackage{soul}
\usepackage{array}
\usepackage{float}
\usepackage{units}
\usepackage{url}
\usepackage{multirow, bigdelim}
\usepackage{amsthm}
\usepackage{amsmath}
\usepackage{amssymb}
\usepackage{xfrac}
\usepackage{stackrel}
\usepackage{graphicx}
\usepackage{setspace}
\usepackage{esint}
\usepackage{makecell}
\usepackage[colorlinks=true,linkcolor=black, citecolor=blue, urlcolor=blue]{hyperref}
\usepackage{epstopdf} 
\usepackage{algorithm}
\usepackage[noend]{algpseudocode}
\usepackage{natbib}
\makeatletter

\providecommand{\tabularnewline}{\\}
\floatstyle{ruled}
\newfloat{algorithm}{tbp}{loa}
\providecommand{\algorithmname}{Algorithm}
\floatname{algorithm}{\protect\algorithmname}

\usepackage[caption=true,font=normalsize,labelfont=sf,textfont=sf,margin=10pt,indention=10pt,singlelinecheck=false]{subfig}
\usepackage{relsize}
\usepackage[table]{xcolor}
\usepackage{diagbox, pict2e}
\usepackage{setspace}
\allowdisplaybreaks

\begin{document}
\begin{frontmatter}
	\title{ Improved image display by identifying the RGB family color space}         % used by \maketitle

\author [UdS]{Elvis Togban\corref{cor1}}
\ead{E.Togban@usherbrooke.ca}

\author [UdS]{Djemel Ziou\corref{cor2}}
\ead{Djemel.Ziou@usherbrooke.ca}
\address[UdS]{ D\'{e}partement d'Informatique, Facult\'{e} des Sciences, 2500 Bl. de l'universit\'{e}, Universit\'{e} de Sherbrooke, J1K2R1, Sherbrooke, Qu\'{e}bec, Canada.}

\begin{abstract}
To display an image, the color space in which the image is encoded is assumed to be known. Unfortunately, this assumption is rarely realistic. In this paper, we propose to identify the color space of a given color image using pixel embedding and the Gaussian process. Five color spaces are supported, namely Adobe RGB, Apple RGB, ColorMatch RGB, ProPhoto RGB and sRGB. The results obtained show that this problem deserves more efforts.
	
\end{abstract}


\end{frontmatter}

\section{Introduction \label{sec:introduction}}
\begin{figure}[!tbh]
 	
 	\centering
 	\captionsetup{width=0.95\columnwidth}
 	\begingroup
 	\captionsetup[subfigure]{justification=centering,position=top,width=0.3\columnwidth}
 	
 	\subfloat[sRGB.]{\protect\includegraphics[width=0.32\columnwidth]{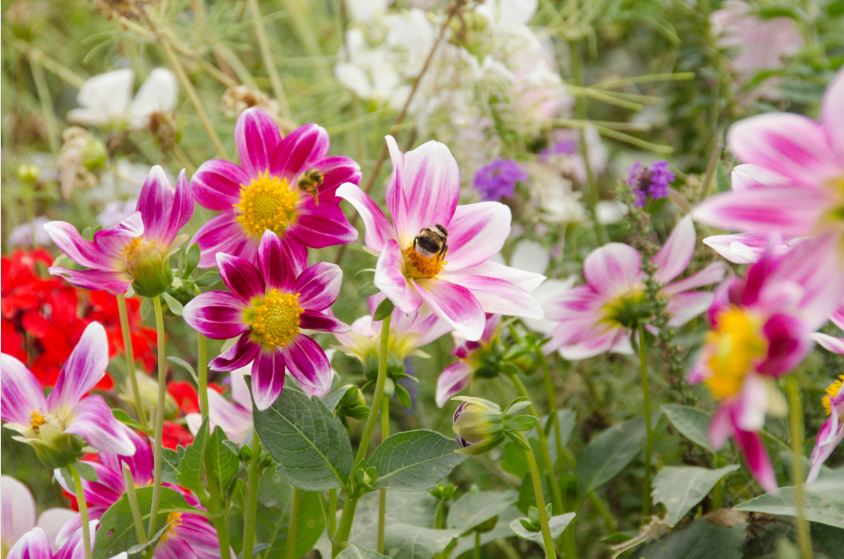}}\hspace{.1cm}\subfloat[Apple RGB.]{\protect\includegraphics[width=0.32\columnwidth]{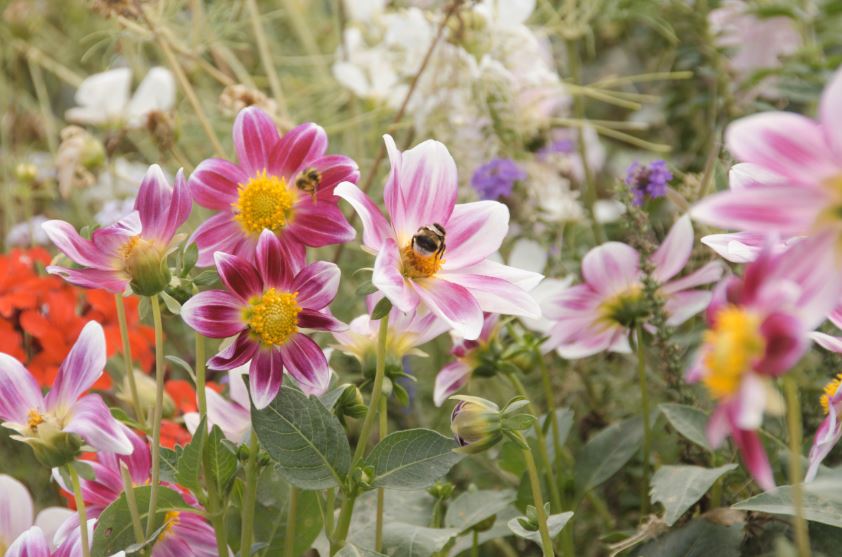}
 	}\subfloat[ProPhoto RGB.]{\protect\includegraphics[width=0.32\columnwidth]{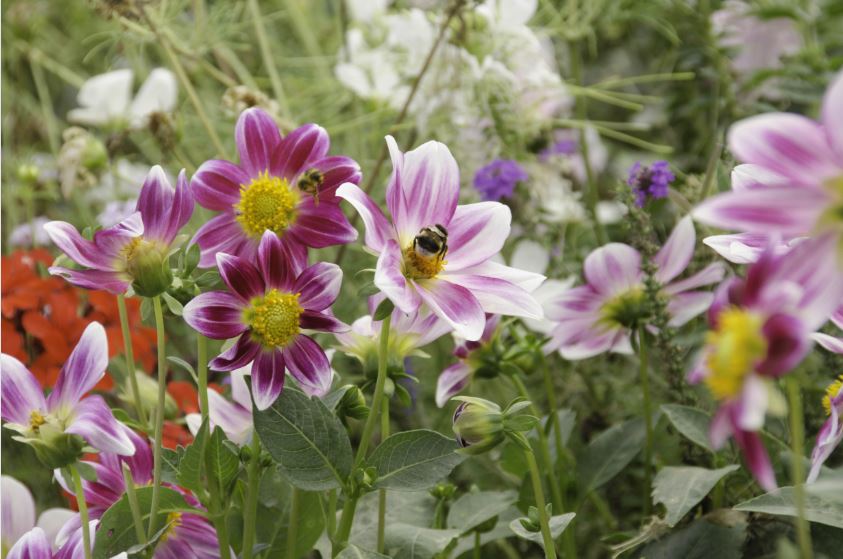}
 	}\hfill{} \\
 	
 	\subfloat{\protect\includegraphics[width=0.32\columnwidth]{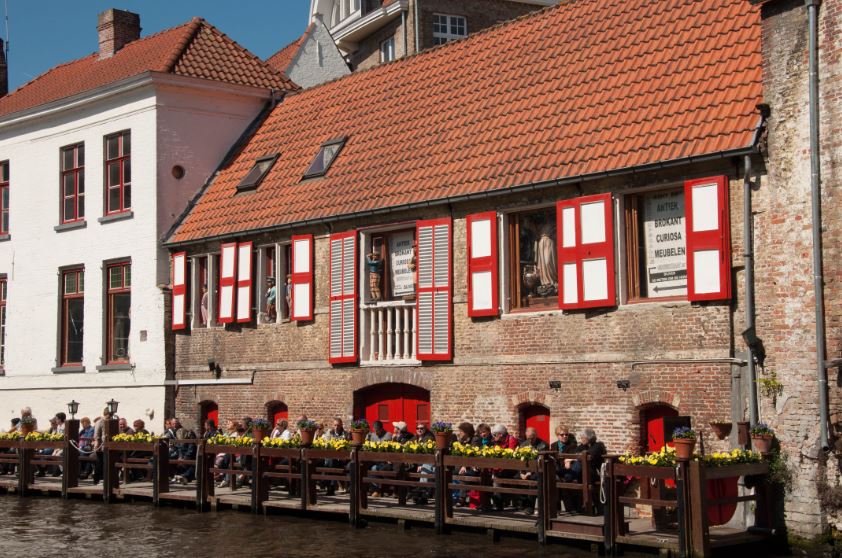}}\hspace{.1cm}\subfloat{\protect\includegraphics[width=0.32\columnwidth]{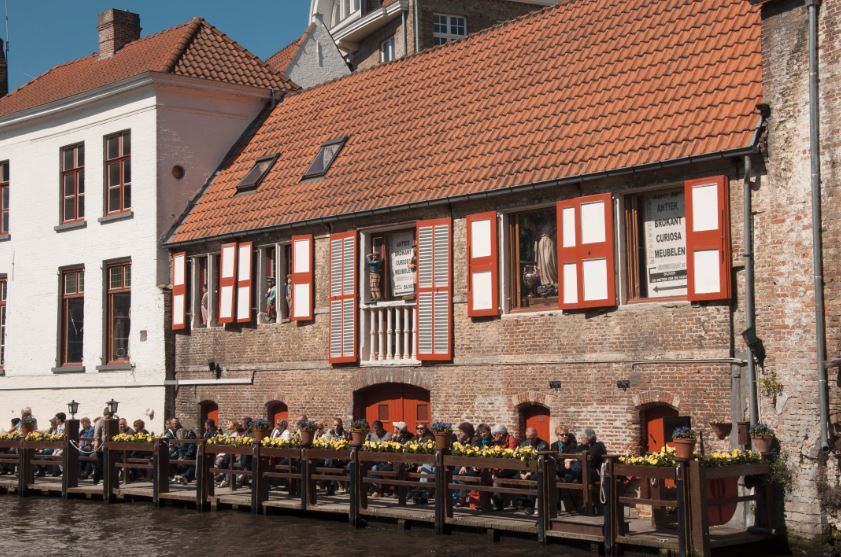}
 	}\subfloat{\protect\includegraphics[width=0.32\columnwidth]{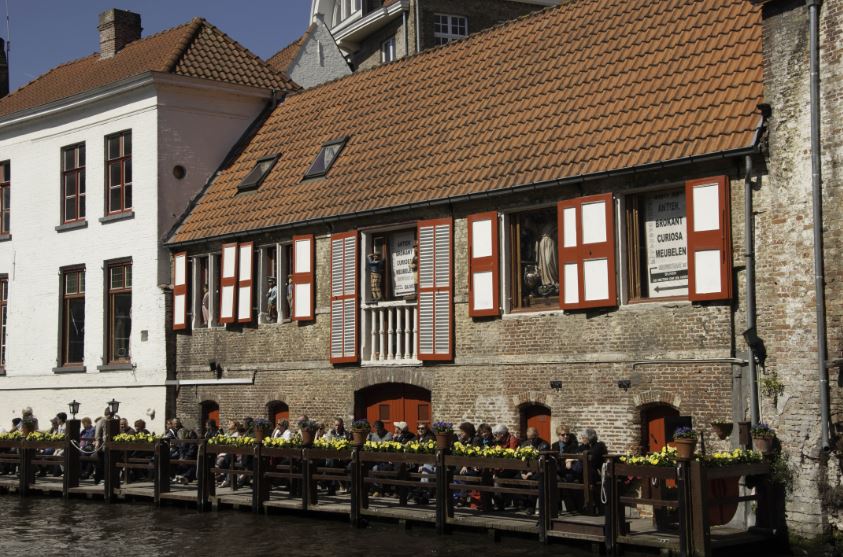}
 	}\hfill{}\\
 	
 	\subfloat{\protect\includegraphics[width=0.32\columnwidth]{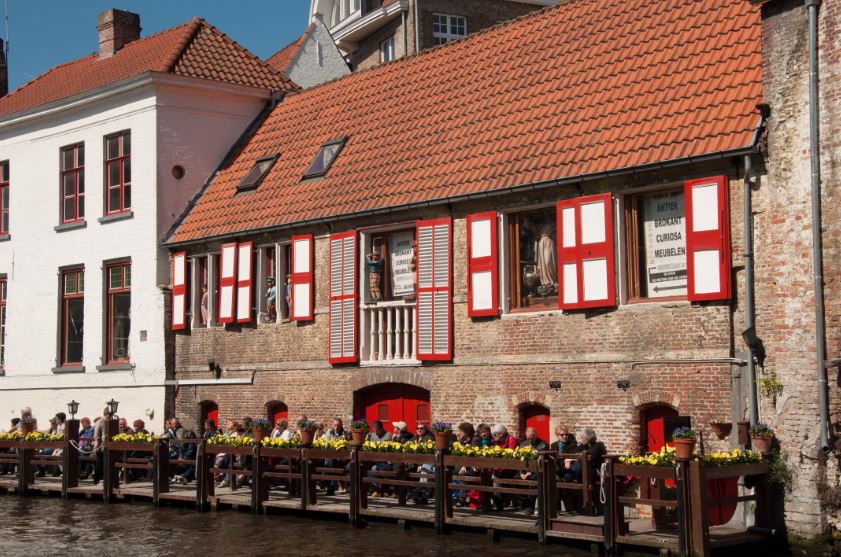}}\hspace{.1cm}\subfloat{\protect\includegraphics[width=0.32\columnwidth]{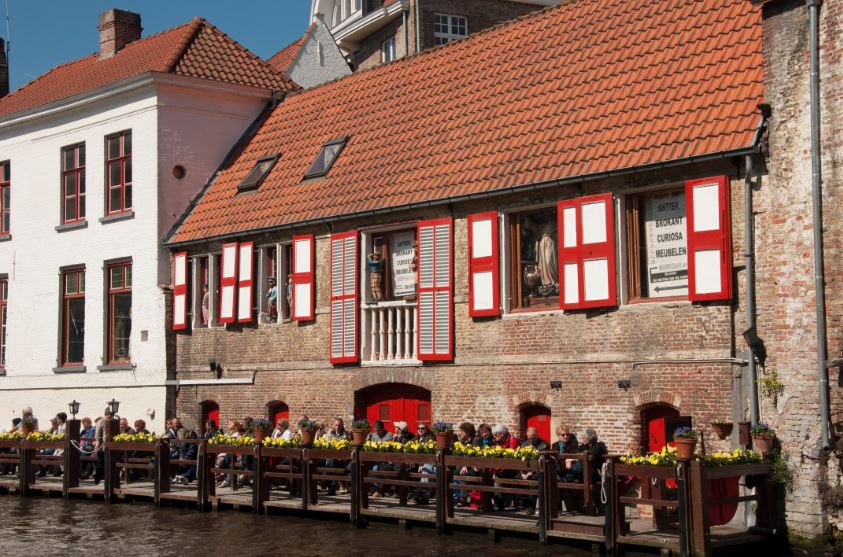}
 	}\subfloat{\protect\includegraphics[width=0.32\columnwidth]{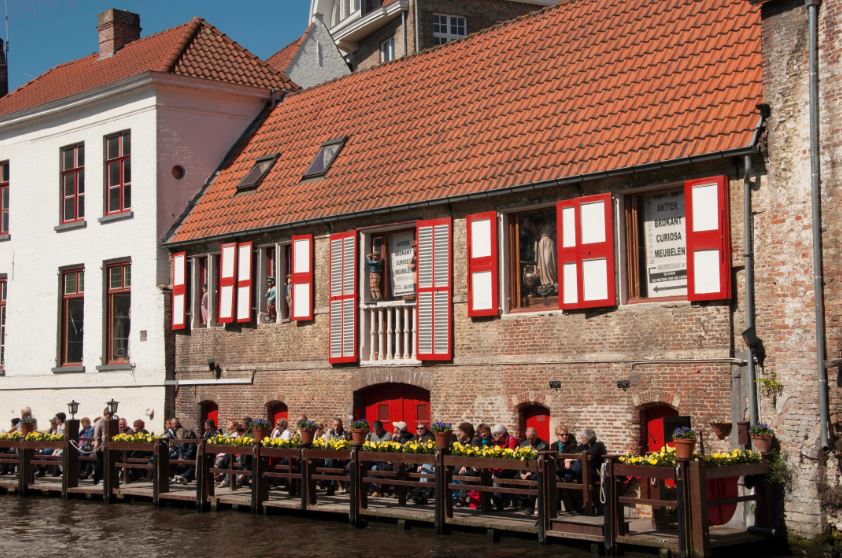}
 	}\hfill{}\\
 	\endgroup
 	\protect\caption{Images displayed with Firefox browser. In the first two rows, the color spaces used for encoding the images are unknown, while those of the images in the last row are known.  \label{fig:color_space}}
 	\vspace{0.15cm}
 \end{figure}

 Colors are the subjective responses linked to psychological and physiological aspects of the human visual system when stimulated to radiation of the spectrum with frequencies that extend from 450 THz  to 750 THz \cite{Distante2020}. Based on  the tristimulus color theory, color space is a color representation system usually based on a set of triplets. RGB color spaces are the most used color space for images display on electronic devices and on the Web. The RGB family  is composed of the channels Red, Green and Blue. In order to faithfully reproduce color in different human activities (displaying, photography, industry), several variants of RGB have emerged such as Adobe RGB, ProPhoto RGB and sRGB.  Moreover, color space can affect computer-based application such as skin detection  \cite{khanam22,filali2012multinomial}, facial age estimation \cite{Hsieh22} and image segmentation \cite{CANALES2024}, white balancing  \cite{kerouh2018content}. Choosing an appropriate color space for a specific visual task is not easy  knowing that an infinite number of color spaces can exist. In some applications or image display systems, the color space in which an image is encoded is assumed to be known. This assumption is rarely realistic. For example, the color space used by default on the Web is sRGB while most professional-level digital cameras use  Adobe RGB. However, after editing an image, it is possible to embed the color space in the metadata. Once this information is available,  the display device faithfully reproduces the colors of the image. However, after various operations on the image, the metadata can be  missing or corrupted which will lead to a misrepresentation of the image. Figure \ref{fig:color_space} shows two images displayed by using the browser Firefox. Browsers consider that the image to be displayed is encoded in the sRGB color space. However, from left to right, each image is encoded respectively in   sRGB, Apple RGB, and ProPhoto RGB. The color space is assumed to be unknown in the first two rows, so there is missing data allowing to carry out the transformation from the space in which the image is encoded to the sRGB space. In other words, regardless of the color space in which the image is encoded, Firefox considers it as encoded in sRGB. As a result, there is a noticeable difference between the three images. In the last row, the color space is known. Therefore, the right transformation is used to convert the images  from the color space in which it is encoded to  sRGB. As a result, the color reproduction is more faithful for the three color spaces in which the image to be displayed is encoded.
 
In this paper, we are interested in the identification of an image color space.  This problem was first addressed  by Vezina et al.  \cite{martin2015}. In their work, they used gamut estimation to discriminate between sRGB, HSV, HLS and Lab color spaces. Unfortunately, in the case of the RGB family, we need a more accurate estimation of the gamut. In this work, we use a feature based on pixels embeding and gaussian process. Five RGB color space namely Adobe RGB, Apple RGB, ColorMatch RGB, ProPhoto RGB and sRGB are considered. The paper is structured as follows. In the next section, we  present our method for color space identification. Section \ref{sec: param_est} is devoted to the model estimation. The experimental results are described in section \ref{expe_resu}.

%\hfill mds

%\hfill August 26, 2015

\section{Color space identification model \label{color_recon}}

RGB color space family can be characterized by the gamut,  the white point, the three primaries (Red, Green, and Blue) and the transfer function. All these attributes could be used to discriminate  color spaces of the RGB family. The gamut and the white point can be estimated from images~\cite{martin2015}. Unfortunately, at the best of knowledge, there is no available algorithm for the estimation of primaries and transfer function from images. Additionally, for the RGB family,  the gamut estimators lack sufficient precision to reliably identify an image's color space. Instead, we propose  pixel embedding and Gaussian process  for color space identification. Let $I$ be a color image encoded in a color space $C_1C_2C_3$ of the RGB family. Basically, the embedding consists in expressing a pixel value as a linear function of its adjacent pixels (in the same channel or in different channel).  More precisely, the pixel embedding can be written as  follows:
\begin{equation}
		I_{k_1}(m,n)=\sum_{(i,j)  \in   V(m, n)}\gamma_{i,j,k_1}I_{k_2}(m+i,n+j)+r_{k_1,k_2}(m,n)
	\label{eq: embed_equa}
\end{equation}
where $\gamma_{i,j,k_1}$  are a set of coefficients such as $\gamma_{0,0,k_1}=0$ if $k_1=k_2$,   $(k_1,k_2) \in \{C_1,C_2,C_3\} \times \{C_1,C_2,C_3\}$ refer to the channels of $C_1C_2C_3$, $I_{k_1}(m,n)$ is the value of the pixel located at $(m,n)$ for the channel $k_1$ and $V(m, n)$ its  neighborhood. A random variable $r_{k_1,k_2}(m,n)$  which is considered normally distributed with zero mean and standard deviation $\sigma_{k_1,k_2}$ is associated with  each location $(n,m)$ and  each pair $(k_1, k_2)$. We propose two possible embeddings  namely the intra-channel embedding ($k_1=k_2$) and the inter-channel embedding ($k_1 \neq k_2$).  The parameters of the embedding model described in Eq.~\ref{eq: embed_equa} are the standard deviation $\sigma_{k_1,k_2}$ and the vector  $\gamma_{k_1}=(\gamma_{i,j,k_1} \mid (i,j)  \in  V(m, n))^t$. The vector is formed by sweeping the neighborhood line by line starting from the top left. In this paper, the neighborhood is  defined  as $V(m,n)= \{(i,j)\mid -J\leq i,j \leq J\}$ where $(2J+1)^2$ is the number of neighboring pixels  of the pixel $(m,n)$.

Consider a collection of labeled images, where the label of an image is the color space in which it is encoded. For each image in this collection, we calculate an embedding vector for each channel in the case of intra-channel embedding and an embedding vector for each pair of channels in the case of inter-channel embedding by using the algorithm~\ref{alg:ar_coeff} described in the next section. The three intra-channel (resp. six inter-channel) embedding vectors are concatenated to form an intra-channel (resp. inter-channel) feature vector. Note that, according to Eq. ~\ref{eq: embed_equa}, the embedding vector $\gamma_{k_1}$ does not depend on the spatial location $(m,n)$ of the pixel. To summarize, the dimension of inter-channel and intra-channel feature vectors are respectively $6(2J+1)^{2}$ and $3((2J+1)^{2}-1)$ per image. The feature vectors are used to train  multinomial logistic regression (MLR)  as a classifier. The choice of MLR is motivated by the fact that it is frugal (less data is needed for training) and its result is easy to explain ~\cite{ksantini2007weighted}. The number of classes is equal to the number of color spaces we need to identify.   For the test, given a color image, the feature vector  is estimated as explained before  and the color space is identified by using the learned classifier.

%\vspace{-2mm}
\section{Parameters estimation \label{sec: param_est}}
\begin{figure*}[!tbh]
	
	\begin{centering}
		
		\subfloat[]{\protect\includegraphics[width=.45\columnwidth]{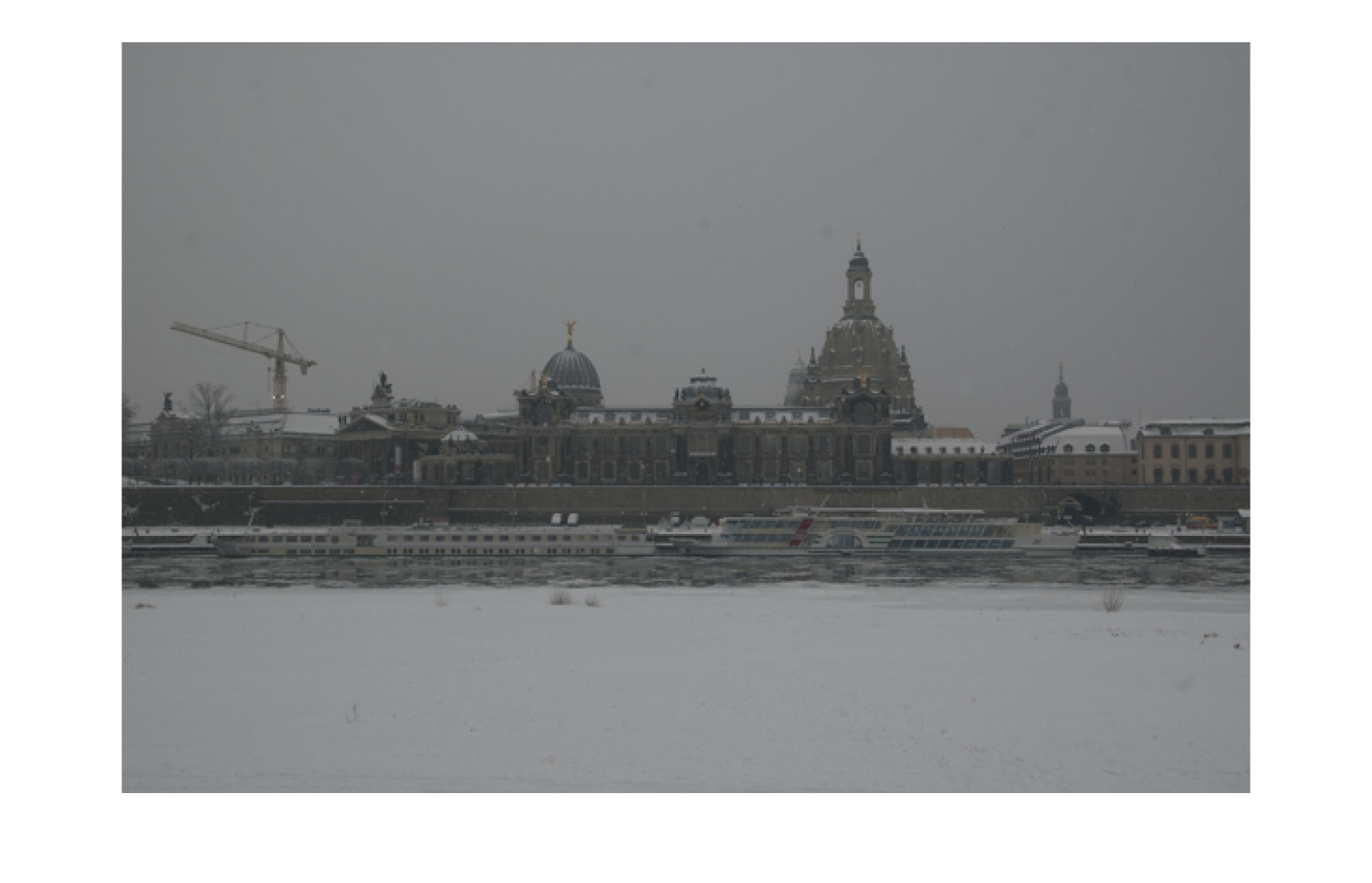} \label{fig:Im_err_ar}} 
		\\

		\subfloat[]{\protect\includegraphics[width=.45\columnwidth]{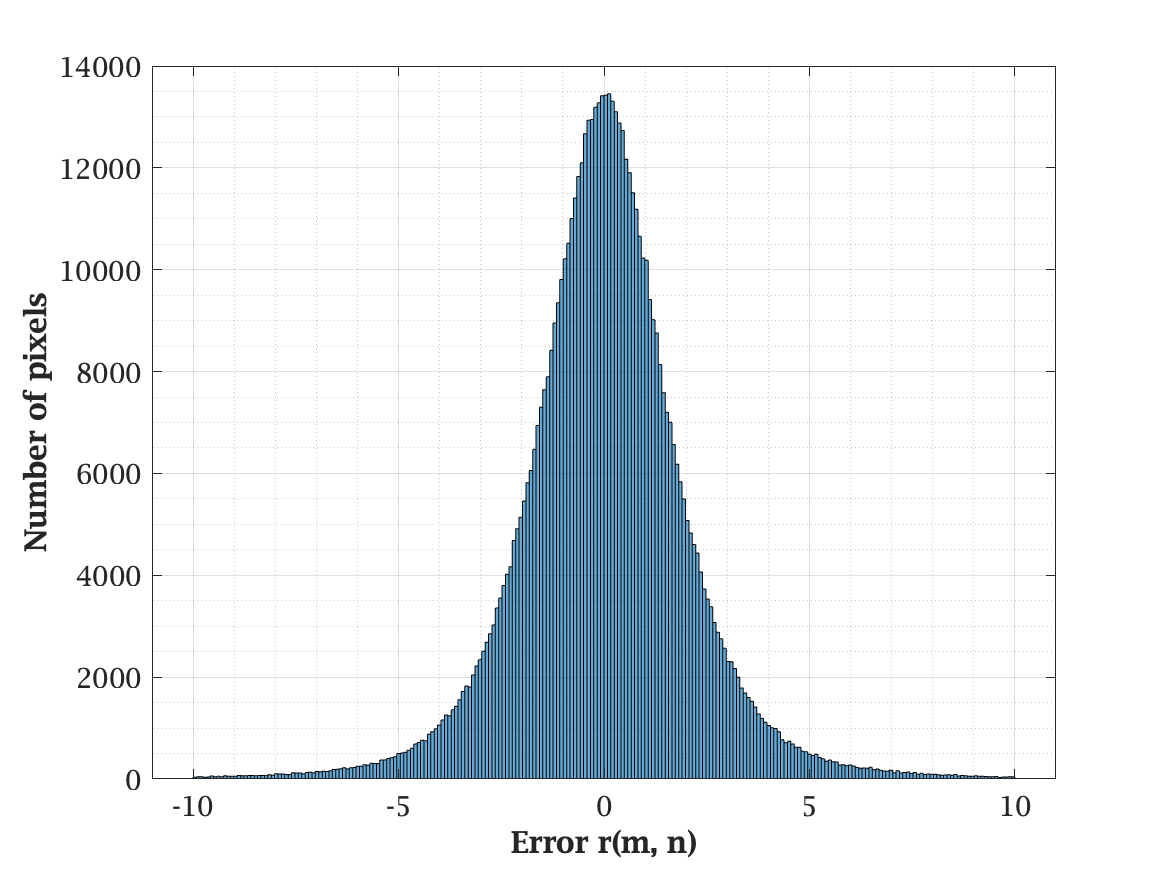} \label{fig:err} }
		\hfill
		\subfloat[]{\protect\includegraphics[width=.45\columnwidth]{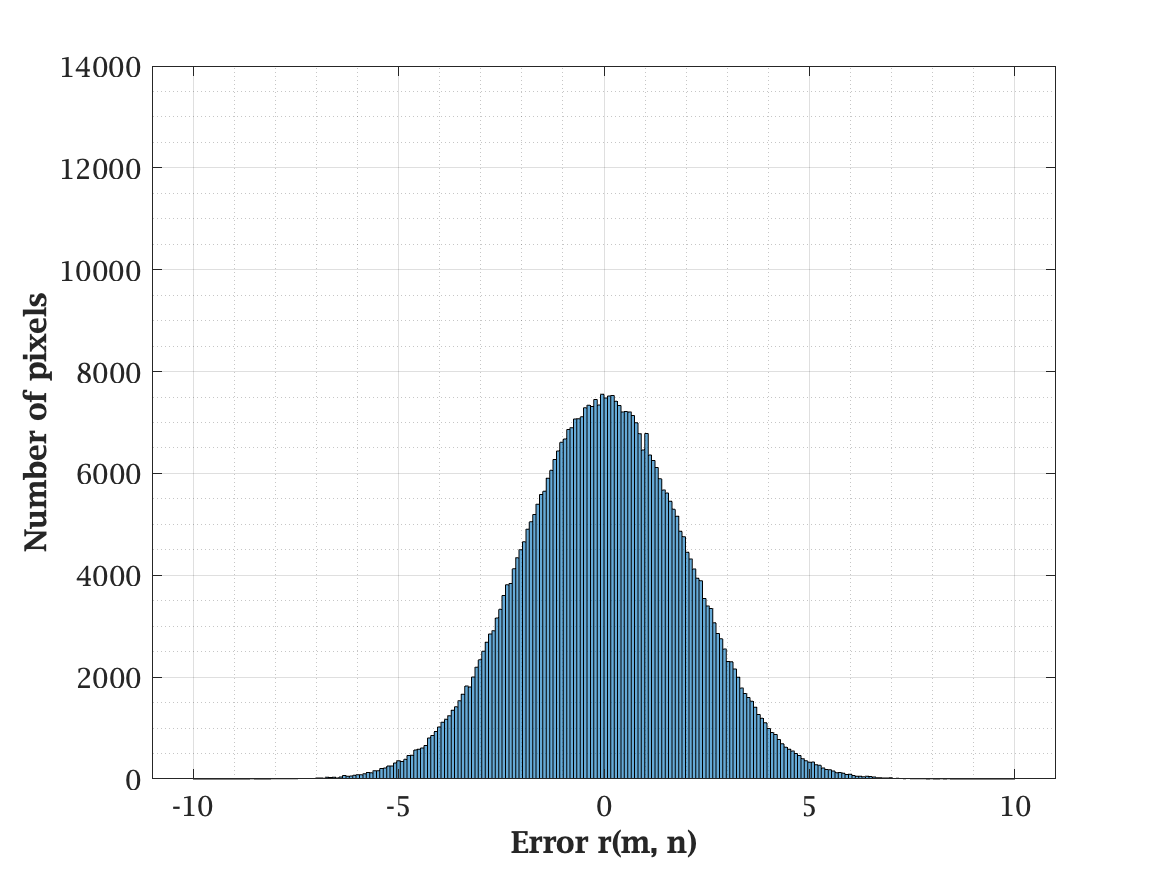}
			
			 \label{fig:gauss_err}}
		
		\protect\caption{ (a) A RGB image. (b) Histogram of the errors $r(m,n)$ computed for the red channel of the image displayed in (a). \label{fig:Proof_err_ar}}
	\end{centering}
	%\vspace{-0.25cm}
\end{figure*}

%\vspace{-2mm}
For the sake of notational simplicity, in the following, $\sigma_{k_1,k_2}$ and $r_{k_1,k_2}()$ will be denoted respectively $\sigma$ and $r()$.
For both inter-channel and intra-channel cases, Eq. \ref{eq: embed_equa} is equivalent to a linear regression model where   $\{I_{k_2}(m+i,n+j) \}_{ (i,j) \; \in \; V(m, n)}$ are the explanatory variables and $I_{k_1}(m,n)$ is the response variable. However, a given pixel may not be linearly correlated to its neighbors. In this case, the  random variable $r(m,n)$ is not  drawn from a Gaussian with zero mean. For the Red channel of a RGB image (fig. \ref{fig:Im_err_ar}), we illustrate the histogram of $r(m,n)$ (fig. \ref{fig:err}). As we can  notice, the errors distribution look like a zero-mean Gaussian. To evaluate the scope of the embedding described in Eq.~\ref{eq: embed_equa}, we estimated the histogram of the variable $r(n,m)$ using a channel or a pair of channels from an image and we performed a K-S Test to decide the normality of this variable $r(n,m)$. This procedure was performed using 1000 pictures of size $602 \times 400$. We noticed that for all the images, the distribution of this error is not a Gaussian. In fact, the error distribution of $r(n,m)$ has a heavier tail and a higher peak than a Gaussian distribution. However, for a sub-sample of pixels,  the error distribution of $r(n,m)$ can be a zero-mean Gaussian (fig. \ref{fig:gauss_err}). For our experiment with 1000 pictures, on average, the error is Gaussian for 36\% pixels in the case of inter-channel and for 30 \% pixels  in the intra-channel case. The equation \ref{eq: embed_equa} is therefore valid for a subset of pixels. 

\captionsetup[figure]{ width=1\textwidth, justification=centering}

Let $I_{k};\; k\in\left\{C_1,C_2,C_3\right\}$ be a channel of $I$ and the pixel $I_k(m, n)$ located at  $(m,n)$ can be  embedded; that means $r_{k_1, k_2}(m,n)$ is a Gaussian variable. Let  
consider the complete data $\{(I_{k_1}(m,n), s(m,n,k_1, k_2))\}_{(m,n) \in V(m,n)}$, where  $s(m,n,k_1, k_2)={s_0(m,n,k_1, k_2),\; \allowbreak s_1(m,n,k_1, k_2)}$ is the set of binary variables. By definition,  $s_0(n,m,k_1,k_2)=1$ (resp. $s_1(n,m,k_1,k_2)=1$) indicates that the pixel  $I_{k_1}(m, n)$ of the channel $k_1$ can (resp. cannot) be embedded over its neighbor of the channel $k_2$. The complete likelihood is given by:
\begin{equation}
 \mathcal{L} =  \prod_{m,n} \prod_{b=0}^1 (\pi_b p(I_{k_1}(n,m)| s_b(n,m,k_1,k_2)=1, \theta_{k_1 b}))^{s_b(n,m,k_1,k_2)} 
 \label{Lik}
\end{equation}
where  $p(s_b(n,m,k_1,k_2)=1)=\pi_b$ is  the  distribution over $s_b(n,m,k_1,k_2)=1$. Let us recall that when $s_0(m,n,k_{1},k_2)=1$,  $r_{k_1,k_2}(m,n)$ is a normal random variable, which leads to consider that  $I_{k_1}(m, n)$ is drawn from a  Gaussian  with mean  $a1(m,n,k_1,k_2) =\underset{(i,j) \in V(m,n)}{\sum}\gamma_{i,j,k_1}I_{k_2}(m+i,n+j)$ and variance $\sigma^2$: 
\begin{equation}
	\begin{split}
		 p(I_{k_1}(m, n)  \mid s_0(m,n,k_{1},k_2)=1, \theta_{k_10})=  
		\mathcal{N} (I_{k_1}(m,n) \mid a1(m,n,k_1,k_2), \sigma^2)
	\end{split}
\end{equation}
where the parameter vector $\theta_{k_10} =(\gamma_{k_1}, \sigma)^t$ are those of the vector $\gamma_{k_1}$, $\sigma$. If $s_1(m,n,k_{1},k_2)=1$, the channel $I_{k_1}(m, n)$ is supposed to be drawn from a Gaussian mixture of $M$ components:  
 \begin{equation}
		 p(I_{k_1}(m, n)  \mid s_1(m,n,k_{1},k_2)=1, \theta_{k_11})=  \sum_{c=1}^M \alpha_c \mathcal{N} (I_{k_1}(m,n) \mid \mu_c, \lambda_c^2)
		 \label{eq: mixture_component}
\end{equation}
where the parameter vector $\theta_{k_11}=(\alpha_1, \mu_1,\lambda_1, \cdots, \alpha_M, \mu_M,\lambda_M)^t$, $\sum_{c=1}^M \alpha_c =1$, and $\alpha_c \ge 0$. 
 We will now  estimate the parameters $\theta_{k_10}$ by using  the  EM algorithm.
 Because the variables $s_b(m,n,k_1,k_2)$ are non-observable, instead of  maximizing the logarithm of the complete-data likelihood  in Eq. \ref{Lik}, we maximize its conditional expectation given by:
\begin{equation}
 \textit{l} =  \sum_{m,n}  \sum_{b=0}^1 \tau_b(m,n,k_1,k_2, \theta_{k_1b}) \ln [\pi_b p(I_{k_1}(n,m)|s_b(m,n,k_1,k_2)=1,\theta_{k_1 b})]
\label{LL}
\end{equation}
where $\tau_b(m,n,k_1,k_2, \theta_{k_1b})$ is the conditional expectation of $s_b(m,n,k_1,k_2)$. The EM algorithm proceeds iteratively in two steps, expectation (E) and maximization (M).   The E-step at the iteration $\eta$,  consists in computing  the posterior $ \tau_b(m,n,k_1,k_2, \theta_{k_1b})$. 
\begin{equation}
	\begin{split}	
    \tau_b(m,n,k_1,k_2, \theta^{(\eta)}_{k_1b})&=p(s_b(m,n,k_1,k_2)=1, \theta^{(\eta)}_{k_1b} \mid I_{k_1}(m, n)) \\
	&=\frac{\pi_b^{(\eta)} p(I_{k_1}(m, n) \mid s_b(m,n,k_1,k_2)=1, \theta^{(\eta)}_{k_1 b}) }{\sum_{b=0}^{1} \pi_b^{(\eta)} p(I_{k_1}(m, n) \mid s_b(m,n,k_1,k_2)=1, \theta^{(\eta)}_{k_1 b}) }
	\end{split}
	\label{tau}
\end{equation}		
The M-Step requires maximizing Eq.~\ref{LL} with respect to $\theta_{{k_1}b}$.  By setting the partial derivative with respect to these variables   to zero and solving we obtain: 
\begin{equation}	
	\pi_b^{(\eta+1)}=\sum_{m,n} \frac{\tau_b(m,n,k_1,k_2, \theta^{(\eta)}_{k_1b})}{N}	\label{pi}
\end{equation}	
where $N$ is the number of pixels.
{\small 
\begin{equation}
	 \sigma^{2(\eta+1)}=  \frac{\underset{m,n}{\sum} \tau_0(m,n,k_1,k_2, \theta^{(\eta)}_{k_1b}) [I_{k_1}(m,n)-\sum_{(i,j)} \gamma^{(\eta)}_{i,j,k_1}I_{k_2}(m+i,n+j)]^2 }{\sum_{m,n} \tau_0(m,n,k_1,k_2, \theta^{(\eta)}_{k_1b})}
  \label{sig}
\end{equation}
}
For the parameter vector $\gamma_{k_1}$, we have to solve the following equation:
\begin{equation}
	A^{(\eta)} \gamma_{k_1}^{(\eta+1)}= Y^{(\eta)} \label{gamma}
\end{equation}
where $A$ is a symmetric matrix with 
{\small
\begin{equation}
A^{(\eta+1)} = 
	\begin{pmatrix}
		a_{-J,-J,-J,-J} & a_{-J,-J,-J,-J+1} & \cdots & a_{-J,-J,J,J} \\
        a_{-J,-J,-J,-J+1} & a_{-J,-J+1,-J,-J+1} & \cdots & a_{J,J,-J,-J+1} \\
        \vdots  & \vdots  & \ddots & \vdots  \\
         a_{-J,-J,-J,J} & a_{-J,-J+1,-J,J} & \cdots & a_{J,J,-J,J} \\
         a_{-J,-J,-J+1,-J} & a_{-J,-J+1,-J+1,-J} & \cdots & a_{J,J,-J+1,-J} \\ 
        \vdots  & \vdots  & \ddots & \vdots  \\   
		a_{-J,-J,J,J} & a_{-J,-J+1,J,J} & \cdots & a_{J,J,J,J} \\
	\end{pmatrix} 
\end{equation}
}
where
\begin{equation}
	a^{(\eta)}_{i,j,u,v}=\sum_{m,n} \tau_0(m,n,k_1,k_2, \theta^{(\eta)}_{k_1b}) I_{k_2}(m+i,n+j)I_{k_2}(m+u,n+v)
\end{equation}
%

%{\small
\begin{equation}
	\begin{array}{c}

	 \gamma_k^{(\eta+1)} = 
	\begin{pmatrix}
		\gamma^{(\eta+1)}_{-J,-J,k_1} \\
		\gamma^{(\eta+1)}_{-J,-J+1,k_1}\\
		\vdots  \\
		\gamma^{(\eta+1)}_{-J,J,k_1} \\
		\gamma^{(\eta+1)}_{-J+1,-J,k_1}\\
		\vdots  \\
		\gamma^{(\eta+1)}_{J,J,k_1} 
	\end{pmatrix} \\ 
	Y^{(\eta)}=  
	\begin{pmatrix}
		\sum_{m,n} \tau_0(m,n,k_1,k_2, \theta^{(\eta)}_{k_1b})\,I_{k_1}(m,n)\,I_{k_2}(m-J,n-J)\\
		\sum_{m,n} \tau_0(m,n,k_1,k_2, \theta^{(\eta)}_{k_1b})\,I_{k_1}(m,n)\,I_{k_2}(m-J,n-J+1)\\
		\vdots  \\
         \sum_{m,n} \tau_0(m,n,k_1,k_2, \theta^{(\eta)}_{k_1b})\,I_{k_1}(m,n)\,I_{k_2}(m-J,n+J)\\
          \sum_{m,n} \tau_0(m,n,k_1,k_2, \theta^{(\eta)}_{k_1b})\,I_{k_1}(m,n)\,I_{k_2}(m-J+1,n-J)\\
          \vdots  \\
		\sum_{m,n} \tau_0(m,n,k_1,k_2, \theta^{(\eta)}_{k_1b})\,I_{k_1}(m,n)\,I_{k_2}(m+J,n+J)
	\end{pmatrix}
	\end{array}
\label{eq:gamma_Y}
\end{equation}
%}
Again we use the EM for the estimation of the parameters $\theta_{k_11}$  of the Gaussian mixture (second level of the EM). Consider the complete data $\{(I_k(m,n), \\ z(m,n))^t\}_{(m,n) \in V(m,n)}$, where a single component of the vector $z(m,n)$ of dimension $M$ is one; i.e. $z(m,n) =(0,\cdots,1,\cdots,0)^t$. A  vector $z(m,n)$ indicates the component from which $I_k(m,n)$ originates; $p(I_k(m,n)|z_c(m,n) =1, \theta_{k_11}) = \mathcal{N} (I_{k_1}(m,n) \mid \mu_c, \lambda_c^2)$ and $p(z_c(m,n) =1)=\alpha_c$. The conditional expectation of the logarithm of the complete-data likelihood  is given by:
\begin{equation}
	\begin{split}
	\textit{l}_2 =  \sum_{m,n}  \tau_1(&m,n,k_1,k_2,  \theta_{k_1b})  \sum_{c=1}^M \iota_c(m,n,k_1, \theta_{k_11}) \ln [\alpha_c p(I_{k_1}(n,m)|\\ z_c(m,n)  &=1,\theta_{k_11})]
	\end{split}
	\label{LLM}
\end{equation}
where $\iota_c(m,n,k_1, \theta_{k_11})$ is the conditional expectation of $z_c(m,n,k_1)$. Note that,  if one center of the Gaussian mixture coincides with a sample observation and the corresponding variance tends to zero, then the likelihood function increases without limit, leading to a degenerate EM estimator. For this, we consider  that all the components of the Gaussian mixture have the same variance; $\lambda_c^2=\lambda^2$.
In the E-step, we estimate the conditional posterior:
 \begin{equation}
	\begin{split}	
	\iota_{c}(m,n,k_1, \theta^{(\eta)}_{k_1 1}) &=p(z_c(m,n) =1, \theta^{(\eta)}_{k_1 1} \mid I_{k_1}(m, n)) \\
	&=\frac{\alpha^{(\eta)}_c p(I_{k_1}(m, n) \mid z_c(m,n) =1,\theta^{(\eta)}_{k_11})  }{\sum_{c=1}^{C} \alpha^{(\eta)}_c p(I_{k_1}(m, n) \mid z_c(m,n) =1, \theta^{(\eta)}_{k_11}) }
	\end{split}
	\label{iota}
\end{equation}		
In the M step, the following updating rules of the mixture parameters  are obtained by maximizing the conditional expectation of the complete-data log-likelihood in Eq.~\ref{LLM}:
\begin{equation}
	\alpha^{(\eta+1)}_{c} = \frac{\sum_{m,n} \tau_1(m,n,k_1,k_2, \theta^{(\eta)}_{k_1b}) \iota_{c}(m,n,k_1, \theta^{(\eta)}_{k_1 1})}{\sum_{m,n} \tau_1(m,n,k_1,k_2, \theta^{(\eta)}_{k_1b})} \label{alpha}
\end{equation}
{\small 
\begin{equation}
	 \lambda^{2(\eta+1)}=  \frac{\underset{m,n}{\sum}  \tau_1(m,n,k_1,k_2, \theta^{(\eta)}_{k_1b}) \sum_{c=1}^M  \iota_{c}(m,n,k_1, \theta^{(\eta)}_{k_1 1})(I_{k_1}(m,n)-\mu^{(\eta)}_c)^2 }{\sum_{m,n} \tau_1(m,n,k_1,k_2, \theta^{(\eta)}_{k_1b})}
  \label{lambda}
\end{equation}
}

 \begin{equation}
	 \mu_c^{(\eta+1)}=  \frac{\underset{m,n}{\sum} \tau_1(m,n,k_1,k_2, \theta^{(\eta)}_{k_1b}) \iota_{c}(m,n,k_1, \theta^{(\eta)}_{k_1 1}) I_{k_1}(m,n) }{\sum_{m,n}   \tau_1(m,n,k_1,k_2, \theta^{(\eta)}_{k_1b})\iota_{c}(m,n,k_1, \theta^{(\eta)}_{k_1 1})}
  \label{moy}
\end{equation}
 
 The parameters estimation procedure is  summarized in algorithm \ref{alg:ar_coeff}.

\begin{algorithm}[!tbh]
	
	\textbf{Input}:  $I_{k_1}$, $I_{k_2}$, and $J$. \newline
	\textbf{Output}: Return the embedding vector $\gamma_{k_1}$.

	1. Initialization: 
	
	Randomly choose initial values of the parameters $\theta_{k_10}$ and use kmeans to initialize $\theta_{k_11}$.

	2. E-step: For each  pixel value $I_{k}(m,n)$, compute  $\tau(m,n)$ (Eq. \ref{tau}) and  $\iota(m,n)$ (Eq. \ref{iota}).

	\vspace{0.25cm}
	
	3. M-step: Estimate the parameters  $\theta_{k_10}$ and $\theta_{k_11}$ by using Eqs.~\ref{pi},~\ref{sig},~\ref{gamma}, ~\ref{alpha},~\ref{lambda},~\ref{moy}.
	
	\vspace{0.25cm}
	
	4. Repeat the steps 2 and 3 until  convergence.
	\protect\caption{ Algorithm for the embedding  \label{alg:ar_coeff}}
	%\vspace{-0.25cm}
\end{algorithm}

\section{Experimental Results \label{expe_resu}}

We collect  1000  images  in RAW format respectively  from  each of DRESDEN \cite{gloe2010dresden} and RAISE \cite{dang2015raise} collections. Then, we convert each image in  five color spaces namely Adobe RGB, Apple RGB, ColorMatch RGB, ProPhoto RGB and sRGB. The number of images obtained is then 10K. Firstly, the embedding vectors are estimated from all images. The neighborhood is chosen by setting $J \in \left\lbrace 1, 2, 3 \right\rbrace $. In order to select the appropriate number of components $M$ (Eq. \ref{eq: mixture_component}), we use the Akaike information criterion (AIC). We vary $M$ from 2 to 8 with step 2. Once the features have been extracted from each image, we use the Generalized discriminant analysis to reduce the dimensions of the features \cite{baudat2000generalized}. We use  them to train a multinomial logistic regression model (MLR). In our earlier work~\cite{color_2023}, we initially assumed that $I_{k_1}(m, n)$ was drawn from a uniform distribution. However, we have found that this distribution is not suitable.  Therefore we compare our new approach to the one proposed in \cite{color_2023}. We also compare our model with the gamut estimation method  \cite{martin2015}. For this purpose, we use 5k pixels per channel and per image (a total of 40 millions pixels) in order to approximate the gamut. In fact, in their work, the authors estimated the gamut by computing the histogram from a collection of images.  We report  the accuracy (in percentage)  based on a five stratified cross-validation in table \ref{tab:Performance-vs-color}. The results reported are the obtained with $J=2$ since it gives the highest accuracy. The following formula is used to compute the accuracy:

\begin{equation}
Accuracy (\%) = \frac{TD \times 100}{N}
\end{equation}
where $TD$ is the number of images with color space correctly detected and $N$ is the total number of images in the test set.

%\captionsetup{width=1\columnwidth}
\begin{table}[!tbh]
	\centering
	%\normal
	\setlength \extrarowheight{2pt} 
	\setlength{\tabcolsep}{2pt}
	%\resizebox*{1\columnwidth}{.95\textheight}{%
		
		\protect\caption{Accuracy obtained for the color space identification problem. \label{tab:Performance-vs-color}}
		\begin{tabular}{cc|c|c|}
			\tabularnewline
			\cline{2-4}
			\multicolumn{1}{ c|  }{}&\makecell{\textbf{Color}\\\textbf{Space}}&\makecell{\textbf{Accuracy by}\\\textbf{ space (\%)}} &\makecell{\textbf{Accuracy (\%)}}
			\tabularnewline
			\hline
			
			\multicolumn{1}{ |c|  }{\makecell{intra-channel}}&\makecell{\small{Adobe}  \\ Apple \\ColorMatch \\ ProPhoto\\ sRGB} &\makecell{ 69 \\ 69.15 \\ 66.6 \\ 68.05  \\ 70.15 }& \textbf{68.59} $\pm$ 1.34
			\tabularnewline
			\hline
			
			\multicolumn{1}{ |c|  }{\makecell{inter-channel \\ RGB}}&\makecell{\small{Adobe}  \\ Apple \\ColorMatch \\ ProPhoto\\ sRGB} &\makecell{ 67.90 \\ 68.30 \\ 66.7 \\ 67  \\ 71.1 }& 68.20 $\pm$ 1.75
			\tabularnewline
			\hline
			
			\multicolumn{1}{ |c|  }{\makecell{intra-channel \\ \cite{color_2023}}}&\makecell{\small{Adobe}  \\ Apple \\ColorMatch \\ ProPhoto\\ sRGB} &\makecell{ 67.55 \\ 63.7 \\ 66.05 \\ 69.7  \\ 64.55 }& 66.31 $\pm$ 1.28
			\tabularnewline
			\hline
			
			\multicolumn{1}{ |c|  }{\makecell{Gamut \\ Estimation \\ \cite{martin2015}}}&\makecell{Adobe  \\ Apple \\ColorMatch \\ ProPhoto\\ sRGB} &\makecell{ 20.65 \\ 21.6  \\ 22.7\\19.25  \\18.85}& 20.61 $\pm$ 1.18
			\tabularnewline
			\hline		
			
		\end{tabular}
		%}
	
	%\vspace{-1.75cm}
	%	\vspace{-0.25cm}
\end{table} 

As stated in section \ref{color_recon}, the gamut estimation has to be more precise in order to discriminate between color space in the RGB family. Therefore, this method performs poorly (20.61\%, see table \ref{tab:Performance-vs-color}). Our method outperforms by 2\% ($\approx 68\%$) the one proposed in \cite{color_2023}. Indeed, the choice of Gaussian mixture to model the pixels that cannot be embedded over their neighbors is more suitable than a simple uniform distribution. It is worth noticing that the estimation of the embedding vector $\gamma_{k_1}$ depend on the model of  the pixels that cannot be embedded over their neighbors (see Eq. \ref{LL} - Eq. \ref{eq:gamma_Y}). According to the results, for both intra-channel and inter-channel embedding, the sRGB color space is the easiest to identify and the ColorMatch color space is the most difficult. The concatenation of the intra and inter channel embedding vectors, do not improve the identification.

\section{Conclusion \label{Conclusion}}
This paper has proposed a model to identify RGB family color space. In order to identify the color space, we rely on pixel embedding and Gaussian process. When the metadata of an image does not have the information about the color space used, our model can be used to identify it. Once the color space  is identified, we can produce a better display.  When using our model, we were able to correctly identify the color space with an accuracy of 68\%. Improvements can be made by choosing a more flexible distribution than the Gaussian one. In future work, we can also investigate if including image  quality measure \cite{tang2024dtsn} can improve the color space detection.

%\section*{Acknowledgment}
%This work is supported by research grants from the Natural Sciences and Engineering Research Council of Canada (NSERC).

\bibliographystyle{elsarticle-num}
\bibliography{Color_jrnl}

\end{document}